\documentclass[runningheads]{llncs}
\usepackage{graphicx}
\usepackage{amsmath,amssymb} 
\usepackage{color}
\usepackage[width=122mm,left=12mm,paperwidth=146mm,height=193mm,top=12mm,paperheight=217mm]{geometry}

\usepackage{cite}
\usepackage{amsmath,amssymb,amsfonts}
\usepackage{algorithmic}
\usepackage{graphicx}
\usepackage{textcomp}

\usepackage[inline]{enumitem}
\usepackage{algorithmic}
\usepackage{graphicx}
\usepackage{xcolor}

\usepackage{witharrows}

\usepackage{url,hyperref,microtype}
\hypersetup{
    colorlinks=true,
    citecolor=blue,
    linkcolor=blue,
    filecolor=magenta,      
    urlcolor=black
}

\usepackage[flushleft]{threeparttable}
\usepackage{tablefootnote}
\usepackage{array}

\newcolumntype{P}[1]{>{\centering\arraybackslash}m{#1}}
\usepackage{multicol}
\usepackage{multirow}
\usepackage{tabulary}
\usepackage{colortbl}
\definecolor{mygray}{gray}{0.90}
\usepackage{makecell}
\usepackage{booktabs}
\usepackage{subcaption}
\usepackage{stmaryrd}
\usepackage{float}
\usepackage{pifont}

\usepackage{arydshln}
\colorlet{mygray}{gray!15!white}

\begin{document}
\pagestyle{headings}
\mainmatter


\title{MUPA\textsuperscript{2}E: Multimodal Unified Perception with Asymmetric Attention for Emotion Assessment}
\titlerunning{MUPA\textsuperscript{2}E: Multimodal Emotion Assessment}
\authorrunning{S. Gkikas et al.}
\author{Stefanos Gkikas \and
Eric Nichols \and
Christian Arzate Cruz \and
Randy Gomez}
\institute{Honda Research Institute Japan, Wako City, Japan\\
\email{\{stefanos.gkikas, e.nichols, christian.arzate, r.gomez\}@jp.honda-ri.com}}

\maketitle

\begin{abstract}

Automatic emotion assessment can benefit from combining neural and behavioral signals, but many multimodal approaches rely on separate, modality-specific feature-extraction pipelines before fusion. This paper presents MUPA\textsuperscript{2}E, a unified perception framework that processes facial video and electroencephalography (EEG) through a single shared asymmetric-attention backbone. Facial video is represented through axis-folded frame tokens, while EEG is processed either as a raw multichannel waveform or projected into the spatial domain for multimodal fusion. The framework is evaluated on the DMER dataset under a stratified subject-independent protocol, comparing unimodal video, unimodal EEG, and fused video--EEG configurations with per-channel and merged EEG projections. Using the original recordings, with shorter trials zero-padded to match the longest duration, merged fusion at stride~$30$ achieves the highest validation performance and a test accuracy of $70.07\%$. Further analysis revealed that recording duration is unevenly distributed across the affective classes, making the padding pattern a potential classification cue. 
Controlling for this factor by cropping all recordings to a common duration of $20$ seconds yielded a test accuracy of $62.71\%$, providing a stricter duration-controlled assessment of the framework in which differences in recording length are removed as a potential classification cue.
These findings demonstrate the feasibility of processing structurally different neural and visual signals within a compact unified architecture while highlighting the importance of controlling duration-related cues in affective datasets.

\end{abstract}

\section{Introduction}

Automatic recognition of human emotional states is a core challenge in affective computing, underpinning systems designed to sense, interpret, and respond to users' affective states \cite{picard_1997}. Emotional experience is most commonly represented along two independent dimensions: valence, reflecting the positive-to-negative quality of an affective state, and arousal, reflecting its level of activation \cite{russell_1980}. Standardized self-report instruments, such as the Positive and Negative Affect Schedule, capture positive and negative affect and have been widely adopted for ground-truth data collection in affective computing studies \cite{watson_clark_1988}. The Self-Assessment Manikin provides a direct measure of valence, arousal, and dominance and has been applied extensively for affective labeling in laboratory settings \cite{bradley_lang_1994}. These self-report methods depend on explicit user participation, motivating the development of automatic recognition systems that passively infer emotional states from physiological and behavioral signals, with applications ranging from human-computer interaction to social robotics platforms that generate emotion-aware co-speech gestures \cite{vazquez_cruz_gkikas_2026} to body-based emotion recognition systems \cite{cruz_gkikas_acii_2026}. 

Physiological biosignals have been studied extensively as passive channels for monitoring stress-related affective responses without relying on explicit self-report \cite{giannakakis_grigoriadis_2022}. Deep learning has driven substantial progress in stress detection from heterogeneous data sources, including physiological signals, facial expressions, speech, gestures, and other behavioral data \cite{kyrou_kompatsiaris_2025}.
Among physiological modalities, EEG provides direct access to central nervous system correlates of emotional states, encoding neural dynamics that peripheral biosignals, such as electrodermal activity or heart rate, cannot capture directly \cite{alarcao_fonseca_2019}. Early approaches represented EEG using hand-crafted spectral features, with differential entropy computed across frequency bands, establishing a widely adopted representation for emotion classification \cite{duan_zhu_2013}. Subsequent work identified critical frequency bands and electrode channels as primary carriers of emotion-relevant information, motivating spatially and spectrally selective architectures \cite{zheng_lu_2015}. Graph convolutional networks advanced this direction by modeling multichannel EEG features as graphs and learning dynamic inter-channel relationships for emotion recognition \cite{song_zheng_2020}. Convolutional transformer architectures have further improved EEG decoding by jointly modeling local temporal structure and global sequential dependencies \cite{song_zheng_2023}. Multimodal affective datasets pairing EEG recordings, peripheral physiological signals, affective stimuli, and subjective ratings have provided important benchmarks for EEG-based emotion recognition research \cite{koelstra_muhl_2012}.

Facial data provide a complementary and non-intrusive channel for affective sensing, with large-scale in-the-wild databases demonstrating the feasibility of recognizing facial expressions, valence, and arousal from unconstrained facial images \cite{mollahosseini_hasani_2019}. Deep learning architectures for facial expression recognition have advanced substantially, with spatiotemporal models learning to capture both spatial appearance and temporal dynamics across sequences of frames \cite{li_deng_2022}. Continuous valence and arousal estimation from facial video in naturalistic conditions has further confirmed the viability of video as a scalable affective sensing modality \cite{kollias_tzirakis_2019}. The face encodes affect-related cues through observable dynamics, including eye activity, mouth movements, head motion, and facial appearance changes, which have been studied for stress and anxiety recognition from video \cite{giannakakis_pediaditis_2017}. Joint analysis of EEG and facial expressions has demonstrated the added value of combining neural and behavioral information for continuous emotion detection \cite{soleymani_asghari_2016}.

EEG and facial video are complementary modalities for emotion recognition: neural signals reflect internal cortical processing, while facial behavior encodes its external expression; combining the two has been shown to provide non-redundant discriminative information \cite{soleymani_asghari_2016}. Feature-level fusion of EEG trials and facial video sequences has improved upon unimodal baselines in multi-class emotion classification \cite{muhammad_hussain_2023}. Multimodal learning frameworks have been extended to affective categories requiring the joint representation of co-occurring emotional components \cite{li_liu_2026}. Recent work combining facial video and physiological biosignals illustrates this pattern, using modality-specific facial and physiological representations before applying fusion strategies \cite{valergaki_nicodemou_2026}. This design introduces choices that are difficult to generalize and prevents both signals from being processed within a unified computational framework.

In this work, we propose MUPA\textsuperscript{2}E, a framework for multimodal emotion assessment that processes facial video and EEG through a single shared asymmetric attention backbone. Three configurations are evaluated within a common architecture: unimodal facial video, unimodal EEG as a raw multichannel waveform, and multimodal fusion of both, all assessed using a stratified, subject-independent evaluation protocol.

\section{Related Work}
\label{related_work}
Automatic emotion recognition from physiological and behavioral signals has been studied across multiple modalities, including EEG recordings, peripheral physiological signals, and facial video paired with subjective affective ratings \cite{koelstra_muhl_2012}. EEG occupies a distinct position among physiological modalities, capturing central nervous system correlates of affective states with high temporal resolution and encoding neural dynamics that peripheral biosignals cannot directly reflect \cite{alarcao_fonseca_2019}. EEG-based emotion recognition pipelines have combined hand-crafted spectral representations with deep neural architectures, with differential entropy computed across the delta, theta, alpha, beta, and gamma frequency bands establishing an early and widely adopted feature extraction approach \cite{duan_zhu_2013}. Deep networks trained on spectral EEG representations further demonstrated the effectiveness of frequency-band and channel-specific information for EEG-based emotion recognition across subjects \cite{zheng_lu_2015}. Graph convolutional networks subsequently improved on static feature approaches by modeling dynamic inter-channel relationships in EEG-based emotion recognition \cite{song_zheng_2020}, while convolutional transformer architectures have further advanced the modeling of local and global structure in EEG decoding \cite{song_zheng_2023}. More recently, multi-scale temporal modeling combined with a dynamic fusion strategy has been employed for EEG-based emotion recognition, on the dataset also used in this work \cite{gkikas_guo_emotion_prai_2026}.

Facial video provides a behavioral complement to EEG-based approaches, encoding affect-related cues through facial muscle activity, eye movements, head motion, and dynamic changes in appearance \cite{giannakakis_pediaditis_2017}. Graph attention networks operating on differential facial action units have further demonstrated that the relational structure between facial muscle activations carries discriminative information for affective state recognition \cite{kassiotis_stressgat_acii_2026}. Visual affect prediction from facial video has also been studied in the wild, with deep architectures that combine convolutional and recurrent layers demonstrating continuous estimation of valence and arousal from spontaneous facial behavior \cite{kollias_tzirakis_2019}. Joint analysis of EEG and facial expressions has further demonstrated the value of combining neural and facial information for continuous emotion detection \cite{soleymani_asghari_2016}, motivating the investigation of architectures that process both channels within a unified framework.

Multimodal fusion of EEG and facial video has been explored to capture complementary aspects of affective responses, motivated by the idea that neural and behavioral signals reflect different components of emotional processing. Feature-level fusion of EEG trials and facial video sequences has produced improvements over unimodal baselines in multi-class emotion classification \cite{muhammad_hussain_2023}, while broader multimodal affect-recognition research has extended learning frameworks to mixed-emotion settings that require the joint representation of co-occurring affective components \cite{li_liu_2026}. 
Existing fusion pipelines apply modality-specific preprocessing and feature extraction before combining representations at the feature or decision level, introducing design choices that are difficult to generalize across tasks and modality combinations. 
A unified architecture that processes EEG and facial video through a shared backbone, without modality-specific inductive biases in the core model, has not been systematically explored for valence-based emotion classification. 
Modality-agnostic architectures have been investigated in related human state assessment tasks, processing facial videos alongside functional near-infrared spectroscopy data within a single transformer framework \cite{gkikas_tsiknakis_painvit_2024}, and recognizing human states from diverse modalities using a single model \cite{gkikas_arzate_eeite_pain_2026, gkikas_workload_acii_2026, gkikas_arzate_pain_icmi_2026}. These studies indicate that a shared backbone can accommodate behavioral and neural inputs without modality-specific feature extractors, although they target pain rather than valence.

\section{Methodology}
\label{sec:methodology}

\subsection{Signal Representation}

The framework maps facial video and EEG recordings into token sequences, which are processed by a shared asymmetric-attention backbone, enabling evaluation of unimodal and multimodal configurations under a common architecture. Each trial consists of multiple stored segments, which are temporally sorted and concatenated into a complete recording; trials with fewer segments than the maximum are zero-padded to a fixed length.

\subsubsection{Facial video}
For a video of $L_v$ retained frames, each a $224\times224$ RGB image, the temporal dimension is folded into the channel dimension, a step referred to as \textit{axis folding}:
\begin{equation}
\mathbf{X}_v \in \mathbb{R}^{B \times H \times W \times 3L_v},
\end{equation}
where $B$ is the batch size and $H=W=224$. Projecting a one-dimensional biosignal into a visual representation prior to classification has been shown to outperform processing the raw waveform directly in stress detection from electrodermal activity \cite{gkikas_eda_stress_prai_2026}. Geometric information is incorporated by encoding each spatial position $\mathbf{p}\in[-1,1]^2$ with Fourier features, using $K=6$ frequency bands and a maximum frequency $f_{\max}=10$. Since the input has $D=2$ spatial axes, the encoding adds $D(2K+1)=26$ positional features per token:
\begin{equation}
\begin{split}
\gamma(\mathbf{p}) = \bigl[&\sin(\pi s_1 \mathbf{p}),\ \cos(\pi s_1 \mathbf{p}),\ \ldots,\\
                            &\sin(\pi s_K \mathbf{p}),\ \cos(\pi s_K \mathbf{p}),\ \mathbf{p}\bigr],
\end{split}
\end{equation}
where $\{s_k\}_{k=1}^{K}$ spans $[1, f_{\max}/2]$. The spatial axes are flattened into $N = H\times W = 50176$ tokens, with data channels and positional features concatenated per token to form the token matrix:
\begin{equation}
\mathbf{T}_v \in \mathbb{R}^{B \times N \times C'_v}, \qquad C'_v = 3L_v + 26.
\end{equation}
The token sequence is partitioned into $S=32$ contiguous spatial segments of length $n_s = N/S = 1568$.

\subsubsection{EEG waveform}
For unimodal EEG processing, the raw multichannel waveform is used directly without hand-crafted feature extraction. All temporal segments of a trial are concatenated along the time axis, yielding:
\begin{equation}
\mathbf{X}_e \in \mathbb{R}^{B \times C_e \times L_e},
\end{equation}
where $C_e$ is the number of selected EEG channels and $L_e$ is the total concatenated waveform length. Each normalized time position $t\in[-1,1]$ is encoded using Fourier features with $K=6$ bands ($D=1$), adding $2K+1=13$ positional features per token. The time axis is flattened into $L_e$ tokens, with waveform channels and positional features concatenated at each step to form the token matrix:
\begin{equation}
\mathbf{T}_e \in \mathbb{R}^{B \times L_e \times C'_e}, \qquad C'_e = C_e + 13.
\end{equation}
The token sequence is partitioned into $S=32$ contiguous temporal segments of length $n_s = \lceil L_e / S \rceil$, with the final segment zero-padded if necessary.

\subsection{Multimodal Channel Fusion}

In the multimodal configuration, the EEG waveform is projected into the 2D spatial domain and fused with the video tensor along the channel dimension, enabling joint processing of both modalities by a shared 2D backbone. Before projection, the EEG waveform is standardized via z-score normalization followed by min-max rescaling to $[0,1]$. In the merged configuration, this normalization is applied jointly over the full $C_e\times L_e$ amplitude grid; in the per-channel configuration, it is applied independently to each channel prior to stacking. Two projection variants are evaluated.

\subsubsection{Merged}
All $C_e$ channels are treated jointly as a $C_e \times L_e$ amplitude grid and projected to a single image via bilinear interpolation:
\begin{equation}
\mathbf{X}_e^{\mathrm{merged}} \in \mathbb{R}^{B \times H \times W \times 1}.
\end{equation}

\subsubsection{Per-channel}
Each channel is projected independently to a $224\times224$ image via bilinear interpolation, and the results are stacked along the channel dimension:
\begin{equation}
\mathbf{X}_e^{\mathrm{per}} \in \mathbb{R}^{B \times H \times W \times C_e}.
\end{equation}

In both cases, the EEG image is concatenated with the axis-folded video tensor along the channel dimension:
\begin{equation}
\mathbf{X}_m = \bigl[\mathbf{X}_v\ \|\ \mathbf{X}_e\bigr]
\in \mathbb{R}^{B \times H \times W \times (3L_v + n_e)},
\end{equation}
where $n_e = 1$ for merged and $n_e = C_e$ for per-channel. The fused tensor is tokenized identically to the unimodal video case: flattened into $N=50176$ spatial tokens, augmented with 26 Fourier positional features, and partitioned into $S=32$ segments of $n_s=1568$ tokens each.

\subsection{Asymmetric Attention}
\label{sec:asymmetric}

All configurations share a common asymmetric-attention backbone of depth~$1$, comprising a single cross-attention block with a feed-forward sublayer, followed by $R=8$ self-attention rounds, each with a feed-forward sublayer. 
A single cross-attention module operating over a shared latent state has previously been employed for physiological signal classification, yielding competitive performance at a low parameter count \cite{gkikas_kyprakis_resp_2025}. 
A latent state $\mathbf{e}^{(0)}\in\mathbb{R}^{d_0}$, with $d_0=128$, is shared across all $S=32$ segments and instantiated at runtime from a set of $M_0=64$ learnable latent vectors $\{\boldsymbol{\ell}_m\}_{m=1}^{M_0}$:
\begin{equation}
\boldsymbol{\ell}_{\mathrm{init}} = \frac{1}{M_0}\sum_{m=1}^{M_0}\boldsymbol{\ell}_m \in \mathbb{R}^{d_0}.
\end{equation}

\subsubsection{Cross-attention}
Each segment state aggregates information exclusively from its corresponding token subset:
\begin{equation}
\mathbf{e}_s = \mathbf{e}^{(0)} +
\mathrm{Attn}\!\bigl(\mathbf{e}^{(0)},\ \tilde{\mathbf{T}}_s\bigr),
\end{equation}
where $\mathbf{e}^{(0)}\in\mathbb{R}^{B\times 1\times d_0}$ provides the queries and $\tilde{\mathbf{T}}_s\in\mathbb{R}^{B\times n_s\times C'}$ provides the keys and values. The operation is asymmetric: the query is a single vector while the key-value side spans $n_s\gg 1$ tokens. All $S$ segments are processed in parallel. Cross-attention uses $8$ heads, each with a head dimension of $16$.

\subsubsection{Self-attention}
After cross-attention, all segment states are stacked to form $\mathbf{E}\in\mathbb{R}^{B\times S\times d_0}$ and processed through $R=8$ self-attention rounds, enabling global information exchange across all segments:
\begin{equation}
\mathbf{E} \leftarrow \mathbf{E} +
\mathrm{Attn}\!\bigl(\mathbf{E},\ \mathbf{E}\bigr).
\end{equation}
Self-attention also uses $8$ heads, each with a head dimension of $16$. All attention and feed-forward sublayers use pre-layer normalization and residual connections, with attention and feed-forward dropout of $0.10$ applied uniformly. After the final round, segment states are averaged over $S$ and passed through a linear classification head to predict binary valence (positive vs.\ negative).

\subsection{Augmentation \& Regularization}

Several augmentation strategies were applied independently to each modality. Video frames were augmented using \textit{TrivialAugment}~\cite{trivialAugment}, \textit{AugMix}~\cite{augmix}, additive noise, center cropping, and spatial masking. A shared random seed was applied across all frames of a video sample to ensure temporal consistency. EEG waveforms were augmented per channel using additive noise and temporal masking, applied prior to 2D projection in the multimodal configuration. Regularization included label smoothing, attention dropout, and feed-forward dropout. Table~\ref{tab:augm_regul_training} summarizes the complete augmentation, regularization, and training configuration used across all experiments.

\begin{table}
\scriptsize
\caption{Augmentation, regularization, and training configuration.}
\label{tab:augm_regul_training}
\centering
\begin{threeparttable}
\begin{tabular}{P{3.3cm} P{4.3cm}}
\toprule
Method/Parameter & Value \\
\midrule
\midrule
\multicolumn{2}{l}{\textit{Video augmentations}} \\
\midrule
\textit{AugMix}          & $p \in [0.20, 0.40]$ \\\hdashline
\textit{TrivialAugment}  & $p \in [0.20, 0.40]$ \\\hdashline
\textit{Center Crop}     & $p \in [0.20, 0.40]$, $200{\times}200$ \\\hdashline
\textit{Noise}           & $p \in [0.20, 0.40]$, $\sigma = 100$ \\\hdashline
\textit{Masking-1}       & $p \in [0.20, 0.80]$, 3 blocks \\\hdashline
\textit{Masking-2}       & $p \in [0.20, 0.40]$, 20 blocks \\
\midrule
\multicolumn{2}{l}{\textit{EEG augmentations}} \\
\midrule
\textit{Add Noise}        & $p \in [0.10, 0.80]$ \\\hdashline
\textit{Temporal Masking} & $p \in [0.10, 0.80]$, size $\in [0.15, 0.30]$ \\
\midrule
\multicolumn{2}{l}{\textit{Regularization}} \\
\midrule
\textit{Label Smoothing} & $0.15$ \\\hdashline
\textit{Att-Dropout}     & $0.10$ \\\hdashline
\textit{FF-Dropout}      & $0.10$ \\
\midrule
\multicolumn{2}{l}{\textit{Training}} \\
\midrule
Optimizer       & \textit{AdamW} \\\hdashline
Learning rate   & $10^{-4}$ \\\hdashline
LR decay        & \textit{cosine} \\\hdashline
Weight decay    & $0.05$ \\\hdashline
Epochs          & $200$ \\\hdashline
Warmup epochs   & $20$ \\\hdashline
Cooldown epochs & $10$ \\\hdashline
Batch size      & $32$ \\
\bottomrule
\end{tabular}
\begin{tablenotes}
\scriptsize
\item \textit{Att-Dropout}/\textit{FF-Dropout}: dropout probability in
attention/feed-forward sublayers \space
\textit{Masking-1/2}: Cutout on video frames, square $32{\times}32$;
value after \textbar\ = number of blocks \space
\textit{Temporal Masking}: applied per EEG channel prior to 2D projection;
size = fraction of total waveform length masked at the beginning, end,
or center of the waveform \space
\textit{Notes}: $x_1$--$x_2$ means we sample $p\sim\mathcal{U}(x_1,x_2)$
per sample and apply the transform with probability $p$.
\end{tablenotes}
\end{threeparttable}
\end{table}

\begin{figure}
\begin{center}
\includegraphics[scale=0.60]{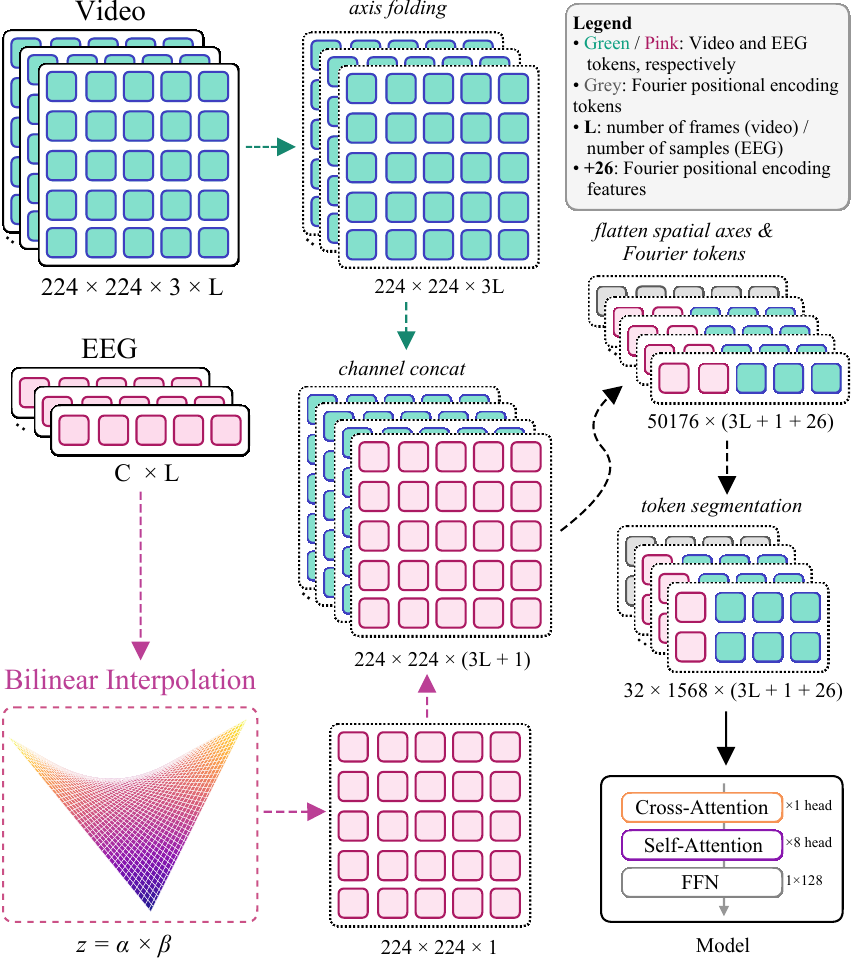}
\end{center}
\caption{Overview of MUPA\textsuperscript{2}E. Facial video is axis-folded into a $224\times224\times3L$ tensor; the EEG waveform ($C\times L$) is projected to a $224\times224\times1$ grid via bilinear interpolation ($z=\alpha\times\beta$). The two representations are channel-concatenated, flattened into $50176$ Fourier-augmented spatial tokens, and segmented into $32$ groups of $1568$ tokens, processed by the asymmetric attention backbone.}
\label{overview}
\end{figure}

\section{Experimental Evaluation \& Results}
This section presents the experimental evaluation of a binary classification task that distinguishes positive from negative affective states. Validation performance is reported using macro-averaged accuracy, precision, and F1 score. Test performance is reported using macro-averaged accuracy.

\subsection{Data Collection}
\label{ssec:data_collection}

This study uses the DMER dataset \cite{yang_liu_2024}, comprising $73$ participants aged $18$ to $35$ (mean $23.06$, SD $3.37$); $7$ of the original $80$ participants were excluded due to physiological signal quality issues. Participants viewed $32$ short video clips drawn from the Stanford film library, selected through a rule-based filtering procedure and expert evaluation to elicit positive, negative, and mixed emotional states: $8$ clips for positive, $8$ for negative, and $16$ for mixed. This work focuses on binary valence classification, distinguishing positive from negative affective states; trials corresponding to mixed emotional content are excluded from all experiments.
EEG was recorded using the DSI-24 wireless dry-electrode system at $300$~Hz across $21$ channels following the international $10$-$20$ system. The time-aligned preprocessed EEG signals provided by DMER are used directly without additional processing; preprocessing steps applied by the dataset authors include independent component analysis, bandpass filtering from $1$ to $50$~Hz, $50$~Hz notch filtering, baseline removal, and re-referencing to Pz with ear electrodes excluded, yielding $18$ channels per trial.
Participants are partitioned into training, validation, and testing sets at the subject level, ensuring no participant appears in more than one split. To avoid performance inflation due to subject-difficulty imbalance, a stratified split protocol is used. Leave-one-subject-out cross-validation is conducted to estimate per-subject classification difficulty, and subjects are ranked by their combined z-score and assigned to four difficulty groups. The final partition comprises $48$ training, $12$ validation, and $13$ testing subjects, with each set drawing from all four groups in proportion. The exact subject-level partition is reported in Table~\ref{tab:subject_split} to support reproducibility and direct comparison with future work.

\begin{table}
\caption{Subject-level split by difficulty group. Subjects are ranked by combined z-score and assigned to four quartile-based groups (Q1\,=\,hardest, Q4\,=\,easiest).}
\label{tab:subject_split}
\begin{center}
\scriptsize
\setlength{\tabcolsep}{3pt}
\begin{threeparttable}
\begin{tabular}{P{1.35cm} P{2.4cm} P{2.4cm} P{2.4cm} P{2.4cm}}
\toprule
\multirow[c]{3}{*}{Split} &
\multicolumn{4}{c}{Difficulty Group} \\
\cmidrule(lr){2-5}
 & Q1 -- Hard & Q2 -- Med-Hard & Q3 -- Med-Easy & Q4 -- Easy \\
\midrule
\midrule
Training (48) &
8, 12, 20, 25, 36, 37, 43, 45, 56, 60, 66, 71 &
14, 30, 33, 35, 39, 50, 51, 53, 55, 65, 72, 73 &
5, 6, 9, 10, 15, 23, 32, 52, 54, 59, 70, 78 &
11, 24, 26, 34, 40, 42, 49, 61, 63, 64, 68, 76 \\\hdashline
Validation (12) &
19, 48, 58 &
22, 41, 79 &
29, 47, 77 &
62, 67, 69 \\\hdashline
Testing (13) &
28, 74, 75, 80 &
21, 38, 57 &
1, 2, 7 &
18, 44, 46 \\
\bottomrule
\end{tabular}
\begin{tablenotes}[para,flushleft]
\scriptsize
\item Q1: $z < -0.36$;\quad Q2: $-0.36 \leq z < -0.08$;\quad Q3: $-0.08 \leq z < 0.17$;\quad Q4: $z \geq 0.17$.
\end{tablenotes}
\end{threeparttable}
\end{center}
\end{table}


\subsection{Unimodal}
\label{sec:unimodal}
Table~\ref{table:unimodal} reports the validation performance of the unimodal video and EEG configurations. For the video modality, the average score varies only slightly across temporal stride settings, spanning $0.62$ percentage points from $68.72$ at stride~$10$ to $69.34$ at stride~$15$. Strides~$20$ and~$30$ achieve average scores of $68.81$ and $69.32$, respectively. Accuracy, precision, and F1 remain closely aligned across all video settings, indicating that no temporal stride produces a clear and consistent advantage. The best video configuration uses a stride of $15$, achieving an average score of $69.34$.
The unimodal EEG configuration achieves the highest unimodal performance, with an average score of $69.81$. This result exceeds the best video configuration by $0.47$ percentage points in average score and by $0.52$ percentage points in accuracy. The EEG accuracy, precision, and F1 scores are $69.79$, $69.87$, and $69.76$, respectively, showing balanced performance across the reported validation metrics.

\begin{table}
\scriptsize
\caption{Unimodal performance across stride settings. \textit{Average}: arithmetic mean of Accuracy, Precision, and F1, used as the primary performance criterion throughout. \textbf{Bold} marks the highest \textit{Average}; \underline{underline} marks the second-highest.}
\label{table:unimodal}
\begin{center}
\begin{threeparttable}
\begin{tabular}{P{1.1cm} P{1.1cm} P{1.2cm} P{1.2cm} P{0.8cm} P{1.2cm}}
\toprule
\multirow[c]{3}{*}{Modality} &
\multirow[c]{3}{*}{Stride} &
\multicolumn{4}{c}{Performance} \\
\cmidrule(lr){3-6}
& & Accuracy & Precision & F1 & \textit{Average} \\
\midrule
\midrule
Video & 10 & 68.75 & 69.05 & 68.36 & \textit{68.72} \\\hdashline
Video & 15 & 69.27 & 69.63 & 69.13 & \underline{\textit{69.34}} \\\hdashline
Video & 20 & 68.75 & 69.05 & 68.63 & \textit{68.81} \\\hdashline
Video & 30 & 69.27 & 69.53 & 69.17 & \textit{69.32} \\\midrule
EEG   & 1  & 69.79 & 69.87 & 69.76 & \textbf{\textit{69.81}} \\
\bottomrule
\end{tabular}
\begin{tablenotes}[para,flushleft]
\scriptsize
\end{tablenotes}
\end{threeparttable}
\end{center}
\end{table}

\subsection{Multimodal}
\label{sec:multimodal}
Table~\ref{table:fusion} reports the validation performance of the multimodal fusion configurations across temporal stride settings. For per-channel fusion, the average score ranges from $68.81$ at stride~$10$ to $69.37$ at stride~$20$, with strides~$15$ and~$30$ yielding $69.32$ and $68.84$, respectively. Performance peaks at stride~$20$ and then declines at stride~$30$. The best per-channel result, with an average score of $69.37$, marginally exceeds the best unimodal video result by $0.03$ percentage points but remains $0.44$ percentage points below the unimodal EEG average of $69.81$.

Merged fusion follows a different pattern. The average scores at strides~$10$, $15$, and~$20$ remain clustered between $68.81$ and $69.34$, consistent with the range observed for per-channel fusion and unimodal video. However, a stride of~$30$ yields a clear improvement. At this setting, merged fusion achieves an accuracy of $70.31$, a precision of $72.68$, and an F1 score of $69.52$, yielding an average score of $70.84$. This is the highest validation result across all unimodal and multimodal configurations. The precision at a merged stride of~$30$ also exceeds accuracy and F1 by a wider margin than in the other fusion settings. Overall, the merged stride~$30$ configuration surpasses the best per-channel result by $1.47$ percentage points in average score and the unimodal EEG configuration by $1.03$ percentage points.

\begin{table}
\scriptsize
\caption{Multimodal fusion performance using Videos and EEG across stride settings.}
\label{table:fusion}
\begin{center}
\begin{threeparttable}
\begin{tabular}{P{1.5cm} P{1.0cm} P{1.2cm} P{1.2cm} P{1.0cm} P{1.2cm}}
\toprule
\multirow[c]{3}{*}{Fusion} &
\multirow[c]{3}{*}{Stride} &
\multicolumn{4}{c}{Performance} \\
\cmidrule(lr){3-6}
& & Accuracy & Precision & F1 & \textit{Average} \\
\midrule
\midrule
Per-channel & 10 & 68.75 & 69.05 & 68.63 & \textit{68.81} \\\hdashline
Per-channel & 15 & 69.27 & 69.53 & 69.17 & \textit{69.32} \\\hdashline
Per-channel & 20 & 69.27 & 69.75 & 69.08 & \underline{\textit{69.37}} \\\hdashline
Per-channel & 30 & 68.85 & 69.05 & 68.63 & \textit{68.84} \\\midrule

Merged  & 10 & 68.75 & 69.05 & 68.63 & \textit{68.81} \\\hdashline
Merged  & 15 & 69.27 & 69.53 & 69.17 & \textit{69.32} \\\hdashline
Merged  & 20 & 69.27 & 69.63 & 69.13 & \textit{69.34} \\\hdashline
Merged  & 30 & 70.31 & 72.68 & 69.52 & \textbf{\textit{70.84}}\\
\bottomrule
\end{tabular}
\begin{tablenotes}[para,flushleft]
\scriptsize
\item Stride refers to the video temporal stride; the EEG waveform stride is fixed at 1 in all multimodal experiments
\end{tablenotes}
\end{threeparttable}
\end{center}
\end{table}

\begin{figure}
\begin{center}
\includegraphics[scale=0.52]{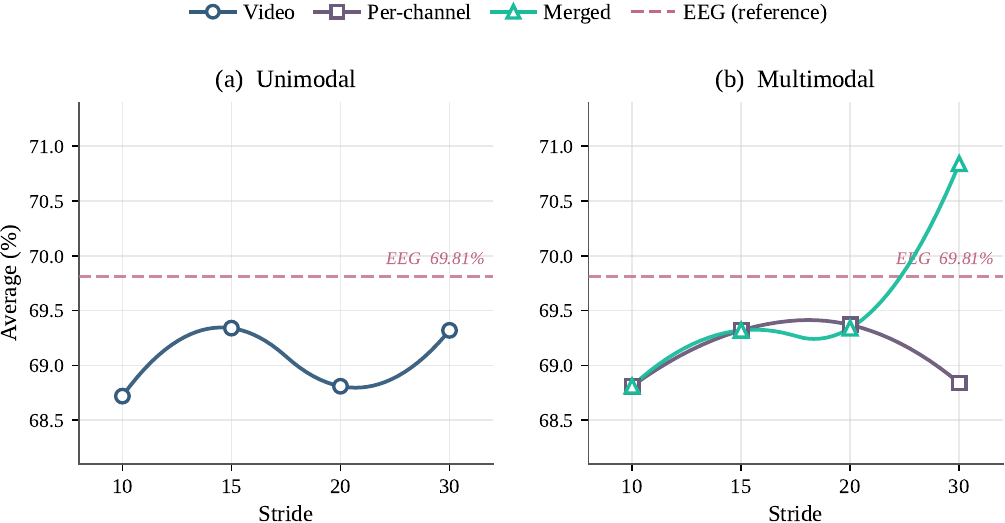}
\end{center}
\caption{Average validation performance (arithmetic mean of Accuracy, Precision, and F1) as a function of video temporal stride for unimodal (a) and multimodal (b) configurations. Unimodal EEG at stride~1 is shown as a dashed reference in both panels.}
\label{comparison}
\end{figure}

\begin{table}
\scriptsize
\caption{Test accuracy, computational cost, and inference efficiency of the best validation-selected merged-fusion configuration.}
\label{table:testing}
\begin{center}
\begin{threeparttable}
\begin{tabular}{P{1.0cm} P{1.0cm} P{1.5cm} P{1.1cm} P{2.3cm} P{2.0cm} P{1.0cm}}
\toprule
\multirow[c]{3}{*}{Fusion} &
\multirow[c]{3}{*}{Stride} &
\multicolumn{2}{c}{Computational Cost} &
\multicolumn{2}{c}{Inference Cost} &
\multirow[c]{3}{1.0cm}{\centering Accuracy} \\
\cmidrule(lr){3-4}\cmidrule(lr){5-6}
& & Params (M) & GFLOPs & Latency (ms) GPU$\downarrow$ & Samples/s GPU$\uparrow$ & \\
\midrule
\midrule
Merged & 30 & 1.89 & 3.13 & 6.55 & 152.65 & \multicolumn{1}{c}{70.07} \\
\bottomrule
\end{tabular}
\begin{tablenotes}[para,flushleft]
\scriptsize
\item Computational and inference cost measured on a single unified sample (Video \& EEG) at inference time on an NVIDIA A100 GPU.
\end{tablenotes}
\end{threeparttable}
\end{center}
\end{table}

\subsection{The Effect of Recording Duration}
\label{ssec:duration}

The experiments reported above initially used the original DMER recordings, preserving their full duration and zero-padding shorter trials as described in Section~\ref{sec:methodology}. After completing these experiments, we examined the distribution of trial durations and identified a potential confounding factor: $62.16\%$ of the positive trials last $30$ seconds, compared with only $25.04\%$ of the negative trials. Consequently, the amount of padding may itself provide class information. Indeed, a simple duration-based rule achieves $68.56\%$ accuracy, close to the $70.07\%$ obtained by the proposed model. We therefore performed an additional controlled experiment using the best validation-selected merged-fusion configuration, cropping all recordings to the first $20$ seconds for both modalities. This follow-up experiment removes recording duration as an available cue while retaining the original evaluation for completeness. Under the same training setup, the cropped configuration achieved $62.71\%$ accuracy, suggesting that recording duration contributed to performance on the original samples.
To further assess subject-independent generalization under the duration-controlled setting, we evaluated the cropped configuration using leave-one-subject-out (LOSO) validation. The framework achieved a mean accuracy of $76.37\%$, providing complementary evidence of its performance across participants.

Finally, \cite{gkikas_guo_emotion_prai_2026} reported $65.22\%$ accuracy using the same stratified hold-out protocol and $20$-second cropped EEG signals, exceeding our cropped multimodal result by $2.51$ percentage points. Their multi-scale temporal windowing strategy may partly explain this difference, consistent with previous findings on physiological-signal analysis \cite{gkikas_kyprakis_resp_2025}.

\begin{table}
\scriptsize
\caption{Accuracy comparison across sample preprocessing and validation protocols.}
\label{table:comparison}
\begin{center}
\begin{threeparttable}
\begin{tabular}{P{1.5cm} P{2.0cm} P{3.0cm} P{1.5cm} P{1.5cm}}
\toprule
Study &Modality &Validation Protocol &Samples &Accuracy \\
\midrule
\midrule

Ours & Video+EEG & Hold-out & Original &70.07 \\\hdashline
Ours & Video+EEG & Hold-out & Cropped  &62.71 \\\hdashline
Ours & Video+EEG & LOSO     & Cropped  &76.37 \\\hdashline

\cite{gkikas_guo_emotion_prai_2026}  &EEG &Hold-out &Cropped &65.22\\

\bottomrule
\end{tabular}
\begin{tablenotes}[para,flushleft]
\scriptsize 
\item Hold-out: is the same stratified hold-out protocol described in this study \space
Cropped: is the same cropped approach of 20 sec described in this study
\end{tablenotes}
\end{threeparttable}
\end{center}
\end{table}

\subsection{Overall Analysis \& Discussion}

Across the unimodal configurations, EEG provides the strongest reference, achieving an average score of $69.81\%$, while video performance remains relatively stable across stride settings. In the multimodal experiments, per-channel fusion remains below the EEG reference, whereas merged fusion at stride~$30$ achieves the highest validation performance, with an average score of $70.84\%$. The corresponding configuration reaches $70.07\%$ accuracy on the held-out test set while remaining compact, with $1.89$M parameters and $3.13$ GFLOPs.

Further analysis showed that recording duration is unevenly distributed between the positive and negative classes, introducing a potential duration-related cue in the original padded samples. When all recordings were cropped to $20$ seconds, and the same merged-fusion configuration was retrained, test accuracy decreased to $62.71\%$. This suggests that recording duration contributed to the performance obtained with the original samples and provides a more controlled estimate when this cue is removed. The cropped configuration also achieved $76.37\%$ mean accuracy under LOSO evaluation, although this result is not directly comparable with the stratified hold-out protocol because of the different training and evaluation partitions.


\section{Conclusion}

This paper presented MUPA\textsuperscript{2}E, a unified perception framework for multimodal emotion assessment in which facial video and EEG are processed through a single shared asymmetric-attention backbone. The framework supports video-only, EEG-only, and fused video--EEG configurations without requiring separate modality-specific feature-extraction backbones.

Under the original padded-sample setting, EEG provides the strongest unimodal performance, while merged multimodal fusion at stride~$30$ achieves the highest validation performance and a test accuracy of $70.07\%$. Additional experiments with all recordings cropped to $20$ seconds reduced the hold-out accuracy to $62.71\%$, suggesting that recording duration contributed to the original performance. These findings demonstrate the feasibility of processing structurally different neural and visual signals within a compact unified architecture while highlighting the importance of controlling duration-related cues in emotion-recognition datasets.

\section*{Acknowledgments}
The authors used large language model (LLM)-based tools for language editing and improvement. All scientific content, results, and conclusions are solely the work of the authors.


\bibliographystyle{splncs}
\bibliography{library}
\end{document}